\documentclass[lettersize,journal]{IEEEtran}
\usepackage{amsmath,amsfonts}
\usepackage{algorithmic}
\usepackage{algorithm}
\usepackage{array}
\usepackage[caption=false,font=normalsize,labelfont=sf,textfont=sf]{subfig}
\usepackage{textcomp}
\usepackage{stfloats}
\usepackage{url}
\usepackage{verbatim}
\usepackage{graphicx}
\usepackage{cite}
\usepackage{multirow}
\usepackage[colorlinks=true,linkcolor=blue,citecolor=blue,urlcolor=blue]{hyperref}
\usepackage{makecell}
\begin{document}

\title{Category-Aware Dynamic Label Assignment with High-Quality Oriented Proposal}

\author{Mingkui Feng, Hancheng Yu$ ^{\ast} $, Xiaoyu Dang$ ^{\ast} $, and Ming Zhou
\thanks{Mingkui Feng, Hancheng Yu, and Ming Zhou are with the College of Electronic and Information Engineering, Nanjing University of Aeronautics and Astronautics, Nanjing 210016, China (e-mail: mkfeng@nuaa.edu.cn; yuhc@nuaa.edu.cn; sz1351419190@nuaa.edu.cn).}
\thanks{Xiaoyu Dang is with Space Information Research Institute, Hangzhou Dianzi University, Hangzhou, 310018, China (e-mail: dang@hdu.edu.cn).}
\thanks{$ ^{\ast} $Corresponding author}}

\markboth{}%
{Shell \MakeLowercase{\textit{et al.}}: A Sample Article Using IEEEtran.cls for IEEE Journals}

\IEEEpubid{}

\maketitle

\begin{abstract}
Objects in aerial images are typically embedded in complex backgrounds and exhibit arbitrary orientations. When employing oriented bounding boxes (OBB) to represent arbitrary oriented objects, the periodicity of angles could lead to discontinuities in label regression values at the boundaries, inducing abrupt fluctuations in the loss function. To address this problem, an OBB representation based on the complex plane is introduced in the oriented detection framework, and a trigonometric loss function is proposed. Moreover, leveraging prior knowledge of complex background environments and significant differences in large objects in aerial images, a conformer RPN head is constructed to predict angle information. The proposed loss function and conformer RPN head jointly generate high-quality oriented proposals. A category-aware dynamic label assignment based on predicted category feedback is proposed to address the limitations of solely relying on IoU for proposal label assignment. This method makes negative sample selection more representative, ensuring consistency between classification and regression features. Experiments were conducted on four realistic oriented detection datasets, and the results demonstrate superior performance in oriented object detection with minimal parameter tuning and time costs. Specifically, mean average precision (mAP) scores of 82.02\%, 71.99\%, 69.87\%, and 98.77\% were achieved on the DOTA-v1.0, DOTA-v1.5, DIOR-R, and HRSC2016 datasets, respectively.

\end{abstract}


\section{Introduction}
\IEEEPARstart{O}{riented} object detection is one of the challenging tasks in computer vision \cite{PIEEE2023_Zou,CVPR2018_DOTA,GRSM2017_Zhu}, which aims to assign a bounding box with a unique semantic category label to each object in the given images \cite{CVPR2018_DOTA, P&RS2020_DIOR, ICPRAM2017_HRSC}. Since these images are often captured from a bird’s-eye view, with objects typically arranged in dense rows in arbitrary directions against a complex background, researchers generally adopted oriented bounding boxes (OBB) to represent oriented objects more compactly. Numerous advanced orientation detectors have evolved by introducing an additional angular branch to traditional horizontal detectors, such as Faster R-CNN-O \cite{TPAMI2017_FasterR-CNNa} and RetinaNet-O \cite{ICCV2017_RetinaNet}, along with various variants that incorporate refined heads \cite{CVPR2019_RoITrans,AAAI2021_R3Det,TGRS2022_S2ANet}. 

\begin{figure*}[!t]
	\centering
	\includegraphics[width=6.5in]{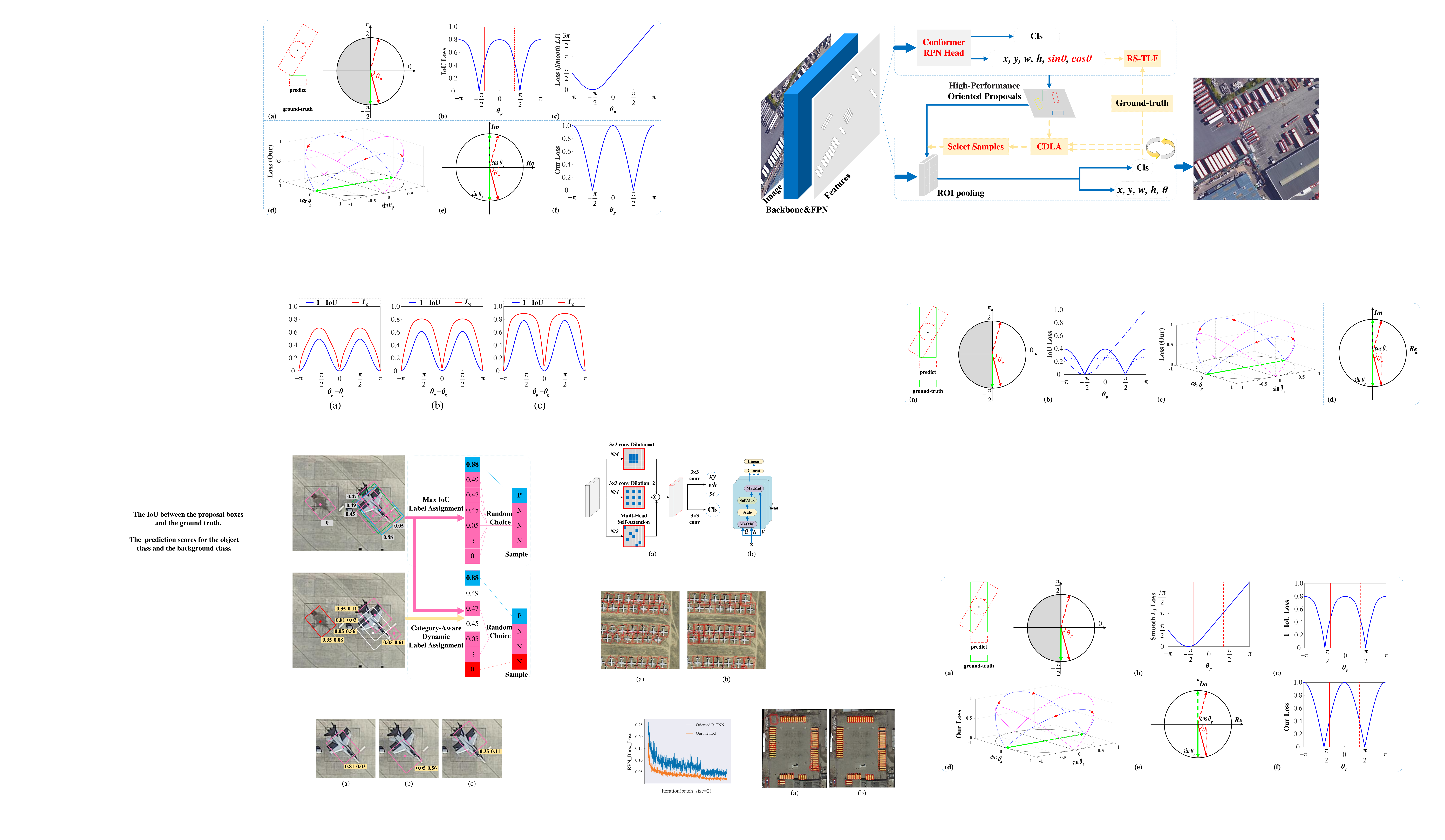}
	\caption{Illustration of the motivation behind the design of the loss function, where arrows indicate direction, $ \theta_{p} $ represents the predicted angle, $ \theta_{g} $ represents the target angle, here $ \theta_{g} $ takes $ -\frac{\pi}{2} $ as an example. (a) Representation and regression of the object OBB angle in the image. (b) The loss curve of $ \text { smooth } L_{1} $ with respect to the predicted angle. (c) The loss curve of 1-IoU with respect to the predicted angle. (d) The relationship between predicted values and our loss function. (e) Representation and regression of the object angle in the complex plane. (f) The loss curve of our loss function with respect to the predicted angle.}
	\label{fig_1}
\end{figure*}

Compared with horizontal detectors, oriented detectors typically need to output the angle of the object's bounding boxes. The angle regression paradigm may potentially have at least two issues. The first issue concerns angle boundary discontinuities, and the second pertains to inconsistencies between the loss function and evaluation metrics. For instance, using the long-side definition method, the OBB angle ranges from $ [-\frac{\pi}{2} ,\frac{\pi}{2} ) $, as shown in Fig. \ref{fig_1}(a). The green line represents the target angle at the angle boundary. The solid red line and dashed red line represent predicted angles near $ -\frac{\pi}{2} $ and $ \frac{\pi}{2} $, respectively. They are equidistant from the target angle. The $ \text { smooth } L_{1} $ loss is illustrated in Fig. \ref{fig_1}(b), where the loss at the red dashed line is much larger than at the red solid line. The $ 1-\text{IoU} $ loss is illustrated in Fig. \ref{fig_1}(c), where the loss at the red dashed line is equivalent to that at the red solid line. It can be seen that the IoU-based loss is an ideal evaluation that can solve the angle boundary discontinuity problem. However, the IoU-based loss for oriented bounding boxes is non-differentiable and cannot be backpropagated.

To address the aforementioned issue, the complex plane coordinates $(\sin \theta, \cos \theta) $ of the boundary-continuous periodic are utilized as replacements for the boundary-discontinuous periodic angle $ \theta $. A smooth and continuous IoU-like loss function is designed as follows:
\begin{equation}\label{eq1}
	L_{\theta}=\left|\sin \theta_{p} \cos \theta_{g}-\cos \theta_{p} \sin \theta_{g}\right|=\left|\sin \left(\theta_{p}-\theta_{g}\right)\right|
\end{equation}
Here, $ \theta_{p} $ represents the predicted angle, and $ \theta_{g} $ represents the target angle. The relationship between the predicted values and the loss function is illustrated in Fig. \ref{fig_1}(d). The blue and magenta curves represent the cases with target angles of $ -\frac{\pi}{2} $ and $ 0 $, respectively. It can be observed from Fig. \ref{fig_1}(d) that every target angle has two convergence points: one is the target angle itself, and the other is its symmetrical point in the complex plane. The loss function guides the predicted angle to converge towards the nearer convergence point. Fig. \ref{fig_1}(e) shows the top view of Fig. \ref{fig_1}(d). Compared with the long-side definition method in Fig. \ref{fig_1}(a), the proposed loss function automatically generates a symmetrical point in the complex plane, solving the boundary discontinuity problem. Fig. \ref{fig_1}(e) shows the side view of Fig. \ref{fig_1}(d). The effectiveness of our loss function is further illustrated by the similarity between Fig. \ref{fig_1}(f) and (b). Compared with the IoU-based loss, our loss function is differentiable.

\IEEEpubidadjcol

The boundary problem of the angle is eliminated by improving the representation and the loss function. However, the regression of angle information is limited by the incompatibility between the classification and regression branches in the orientation detector. Additionally, there is a misalignment between the receptive field of the convolutional features and the OBB \cite{TGRS2022_SoftLabel}, with different object types requiring different receptive fields \cite{ICCV2023_LSKNet}. A conformer RPN head is designed using the strong a priori knowledge inherent in remote sensing images, combined with vanilla convolution and multi-head self-attention mechanisms. This design can dynamically adjust the receptive fields, thereby mitigating the misclassification problem caused by the fixed receptive fields of the detector head and efficiently learning angle information.

The conformer RPN head and the trigonometric loss function generate high-quality proposals for the detector. Traditional IoU-based label assignment often overlooks the impact of the actual shape characteristics in proposals. It is particularly unreasonable to use the same preset IoU threshold for label assignment in oriented object detection. To enhance the rationality of label assignment, a category-aware dynamic label assignment method has been developed based on feedback category prediction information.

In summary, the contributions of this work are:
\begin{enumerate}
	\item{A new angular loss function is proposed by modeling angles in the complex plane, which fundamentally addresses the boundary problem. The method is flexible and simple enough to be integrated into any orientation detection framework.}
	\item{A conformer RPN head that dynamically adjusts the reception field is proposed to effectively correlate classification and regression tasks.}
	\item{Based on feedback category prediction, a category-aware dynamic label assignment method is designed to correct the irrationality of label assignment.}
\end{enumerate}

\section{RELATED WORK}
With the advancement of deep learning, object detection has made significant progress in recent years. This section will first discuss the existing orientation detectors for both HBB and OBB based detection. Then, representative work on OBB representation, loss functions, and label assignment strategy will be presented.

\subsection{Oriented Object Detection}
Early works on oriented object detection were generally derived from horizontal detectors. Due to objects in aerial images typically appearing at arbitrary orientations, they are affected by spatial misalignment between oriented objects and horizontal proposals. Liu \cite{GRSL2016_Liua} and Ma \cite{TOM2018_Maa} address this issue by using rotating anchors with different angles, scales, and aspect ratios. Ding et al. \cite{CVPR2019_RoITrans} proposed a rotating region of interest (RRoI) learner that learns the parameters of the transformation of a horizontal region of interest (RoI) to RRoI under the supervision of OBB. Yang et al. \cite{AAAI2021_R3Det} presented a refined box with aligned features by reconstructing the feature map. In contrast to Yang et al \cite{AAAI2021_R3Det}, $ \text{S}^{2}\text{A-Net} $ \cite{TGRS2022_S2ANet} generates orientation detection results by cleverly using deformable convolution to self-adaptively align deep features. However, these methods require complex RPN operations and extensive boundary box transformation computations. Additionally, some approaches \cite{CVPR2021_ReDet, ICCV2023_LSKNet, CVPR2024_PKINet} has focused on developing high-quality object detectors by enhancing the backbone network. While the aforementioned methods contribute to improving detection performance, they still encounter challenges such as the boundary problem and the label assignment problem.

\subsection{Different Representations for Oriented bounding Boxes}
Recent work has typically focused on developing novel OBB representations to alleviate boundary problems. He et al. \cite{TIP2018_Hea} and Liao et al. \cite{TIP2018_TextBoxesplusplus} used the corner points of oriented rectangles to represent bounding boxes. Gliding Vertex \cite{TPAMI2021_GlidingVertex} introduced four length ratios to form a new representation. Oriented R-CNN \cite{ICCV2021_ORCNN} used parallelogram midpoint offsets to simplify the representation to six-parameter. Yao et al. \cite{TGRS2023_QPDet} designed a five-parameter representation using the circumscribed circle of a rectangular box. Despite avoiding directly using the angles from oriented bounding boxes, these methods require complex transformation calculations and post-processing, such as regularization. In addition to anchor-based methods, anchor-free and keypoint-based methods have also emerged. Chen et al. \cite{ECCV2020_PIoULoss} applied the IoU loss function for OBB based on pixel statistics to anchor-free methods. Oriented RepPoints \cite{CVPR2022_OrientedRepPoints} use adaptive point sets to capture the geometric structure of oriented objects. OSKDet \cite{CVPR2022_OSKDet} explores a novel disordered keypoint heatmap fusion method to learn the shape and direction of the oriented object. Cheng et al.\cite{TGRS2022_AOPG} designed AOPG to generate high-quality proposals. However, these methods require longer training costs to achieve the same results as frame-based methods, and there is semantic ambiguity in the object center point.

\begin{figure*}[!t]
	\centering
	\includegraphics[width=7.0in]{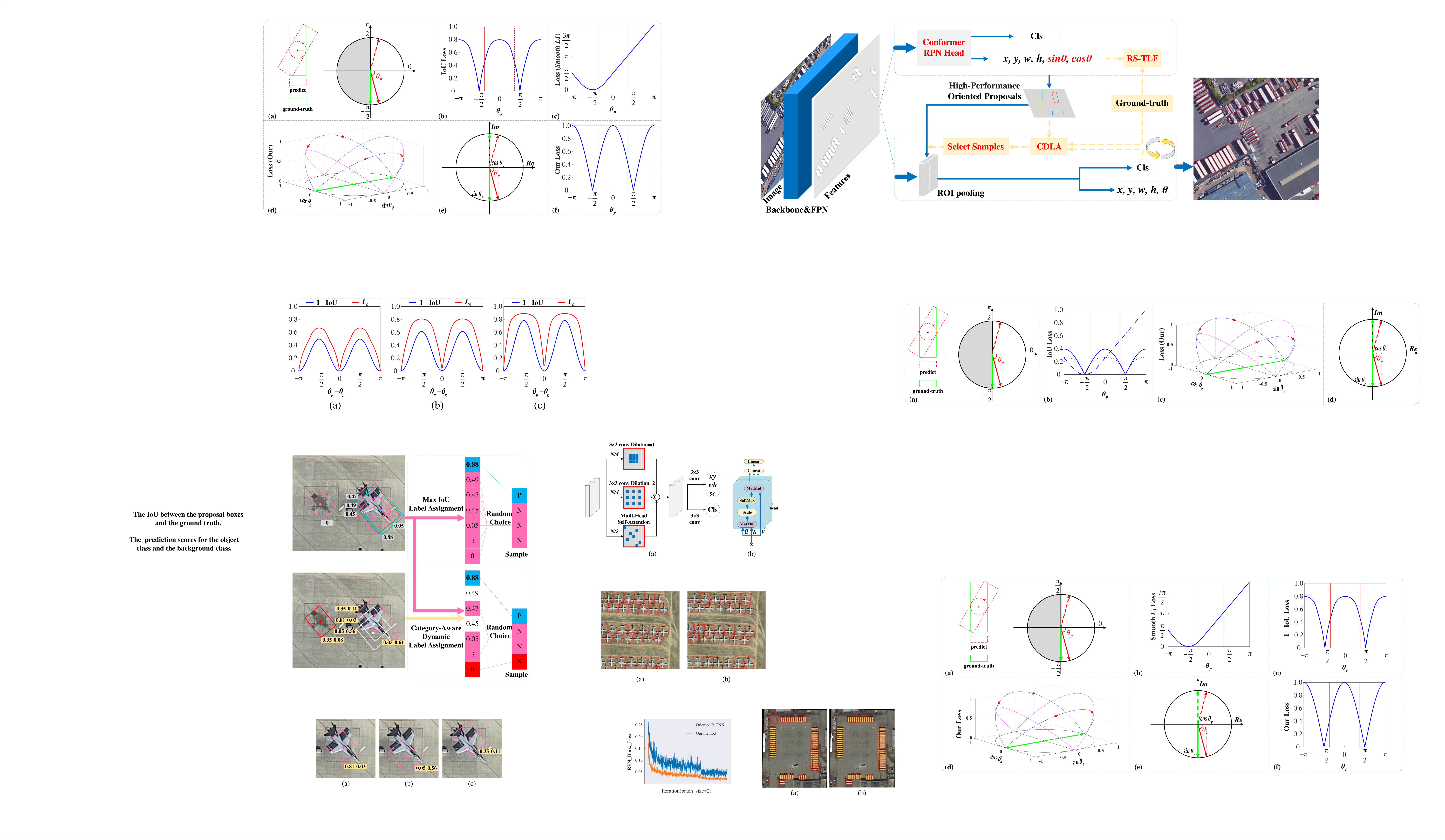}
	\caption{Overall architecture of the proposed approach. The yellow dashed arrows indicate steps that occur only during training, while the blue solid arrows indicate steps shared between training and testing.}
	\label{fig_2}
\end{figure*}

\subsection{Loss Functions}
In addition to using different OBB representation methods, some work attempts to address angle boundary problem through loss functions. Qian et al. \cite{AAAI2021_RSDet} proposed a modulated rotation loss in RSDet to dismiss loss discontinuities and discussed regression inconsistency. However, RSDet \cite{AAAI2021_RSDet} is only a remedial measure taken after the problem is discovered, rather than designing a prediction method that avoids boundary issues. Yang et al. advanced works \cite{ECCV2020_CSL, CVPR2021_DCL, IJCV2022_Yang}, proposed to convert the angle regression problem to a classification task and use circular smooth labels or Densely Coded Labels to replace the one-hot labels, get rid of the boundary discontinuity issue. Wang et al. proposed Gaussian Focal Loss \cite{TGRS2022_GaussianFocalLossa} as a more effective alternative to the classification-based rotation detectors. Yang et al. \cite{TPAMI2023_SCRDetplusplus} design a novel IoU constant factor to alleviate the boundary problem, GWD \cite{ICML2021_GWD} and KLD \cite{NeurIPS2021_KLD} transform an OBB into a 2-D Gaussian distribution and solve the derivation of RIoU problem by computing the distance between the distributions. But the price to pay is the presence of theoretical errors and the increased number of model parameters and calculations.

\subsection{Label Assignment Strategy}
Label assignment is a critical component in selecting high-quality samples for the detector. The sampling issue for horizontal detectors has been more extensively researched. For example, focal loss \cite{ICCV2017_RetinaNet} weights the samples using a reconstruction loss function. ATSS \cite{CVPR2020_ATSS} automatically classifies positive and negative training samples based on the statistical properties of the  object. OTA \cite{CVPR2021_OTA} formulates the sample allocation process as an optimal transmission problem. OHEM \cite{CVPR2016_OHEM} mines some hard example samples for training. PISA \cite{CVPR2020_PISA} assessed the sample difficulty based on mean average precision (mAP). However, there are fewer studies on region sample selection for oriented detectors. Hou \cite{AAAI2022_SASM} added a shape-adaptive strategy to the label assignment. Huang \cite{TIP2022_GGHL} proposed a Gaussian heat map label assignment method. Yu \cite{TGRS2022_SoftLabel} proposed a soft label assignment method. Although these methods have designed dynamic label assignment thresholds, they lack a connection between the detection results and the label assignment.

\section{METHODS}
\subsection{Overview}
This section introduces an improved two-stage oriented object detection framework, as illustrated in Fig. \ref{fig_2}. In the RPN stage, the trigonometric loss function and a conformer RPN head are proposed to generate high-quality oriented proposals. In the ROI stage, a category-aware dynamic label assignment is used to rationally select negative samples. To streamline the discussion related to Fig. \ref{fig_2}, the recommended approach will be elaborated on in the subsequent sections.

\subsection{Trigonometric Loss Function (TLF)}
According to the analysis in the introduction, the OBB is represented using a six-element tuple $ (x, y, w, h, \sin \theta, \cos \theta) $. Here $ (x, y) $ are the coordinates of the center, $ w $ is the length of the long side, $ h $ is the length of the short side, and $ \theta $ is the angle between $ w $ and the x-axis. The OBB regression loss function for RPN training is designed as follows:

\begin{equation}\label{eq2}
	\hspace{-1mm}
	\boldsymbol{L}_{\mathrm{reg}}(\boldsymbol{t}, \boldsymbol{t^{p}})=\sum_{i \in\{x, y, h, w \}} \text { smooth } L_{1} \left(t_{i}-t_{i}^{p}\right) + L{(t_{\theta}, t_{\theta}^{p})}
\end{equation}
where the parameters $ \boldsymbol{t}=(t_{x},~ t_{y},~ t_{w},~ t_{h},~ t_{\sin \theta},~ t_{\cos \theta}) $ and $ \boldsymbol{t^{p}}=(t_{x}^{p},~ t_{y}^{p},~ t_{w}^{p},~ t_{h}^{p},~ t_{\sin \theta}^{p},~ t_{\cos \theta}^{p}) $ denote the offsets of the ground truth and predicted proposal relative to the anchor box, respectively. The variables $ (x, y, w, h) $ use $ \text { smooth } L_{1} $ loss, while $ \theta $ use the $ L{(t_{\theta}, t_{\theta}^{p})} $ loss as defined in equation (\ref{eq5}).

The offsets $ \boldsymbol{t} $ are calculated from the ground truth and anchor box. The encoding equation is as follows:
\begin{equation}\label{eq3}
	\begin{cases}
		t_{x} = ((x_{g} - x_{a}) \cos \theta _{a} + (y_{g} - y_{a}) \sin \theta _{a}) / w_{a} \\ 
		t_{y} = (-(x_{g} - x_{a}) \sin \theta _{a} + (y_{g} - y_{a}) \cos \theta _{a}) / h_{a} \\ 
		t_{w} = \ln_{}{(w_{g} / w_{a})} \\ 
		t_{h} = \ln_{}{(h_{g} / h_{a})} \\  
		t_{\sin \theta} = (\sin \theta _{g} \cos \theta _{a} - \cos \theta _{g} \sin \theta _{a}) \\
		t_{\cos \theta} = (\cos \theta _{g} \cos \theta _{a} + \sin \theta _{g} \sin \theta _{a})
	\end{cases}
\end{equation}

The proposals are calculated based on the offsets $ \boldsymbol{t^{p}} $ and anchor boxes. This decoding equation is as follows:
\begin{align}\label{eq4}
	\hspace{-20mm}
	\begin{cases}
		x_{p} = t_{x}^{p} w_{a} \cos \theta _{a} - t_{y}^{p} h_{a} \sin \theta _{a} + x_{a} \\ 
		y_{p} = t_{x}^{p} w_{a} \sin \theta _{a} + t_{y}^{p} h_{a} \cos \theta _{a} + y_{a} \\ 
		w_{p} = w_{a} e^{t_{w}^{p}} \\ 
		h_{p} = h_{a} e^{t_{h}^{p}} \\ 	
		\sin \theta _{p} = (t_{\sin \theta}^{p} \cos \theta_{a} + t_{\cos \theta}^{p} \sin \theta_{a}) \\ 
		\cos \theta _{p} = (t_{\cos \theta}^{p} \cos \theta_{a} - t_{\sin \theta}^{p} \sin \theta_{a})
	\end{cases}
\end{align}

During the encoding and decoding process, it is common to swap the width and height when the angle between the prior and the target exceeds $ \frac{\pi}{4} $. Although this reduces the extreme values of the angle loss, it relies on the precise prediction of the width and height. The application of trigonometric functions in the encoding, decoding, and loss functions completely resolves the angle boundary problem. Consequently, our encoding and decoding process has removed the swap width and height operation, effectively preventing the influence between the model's angle predictions and width and height predictions.

\begin{figure}[!t]
	\centering
	\includegraphics[width=3.5in]{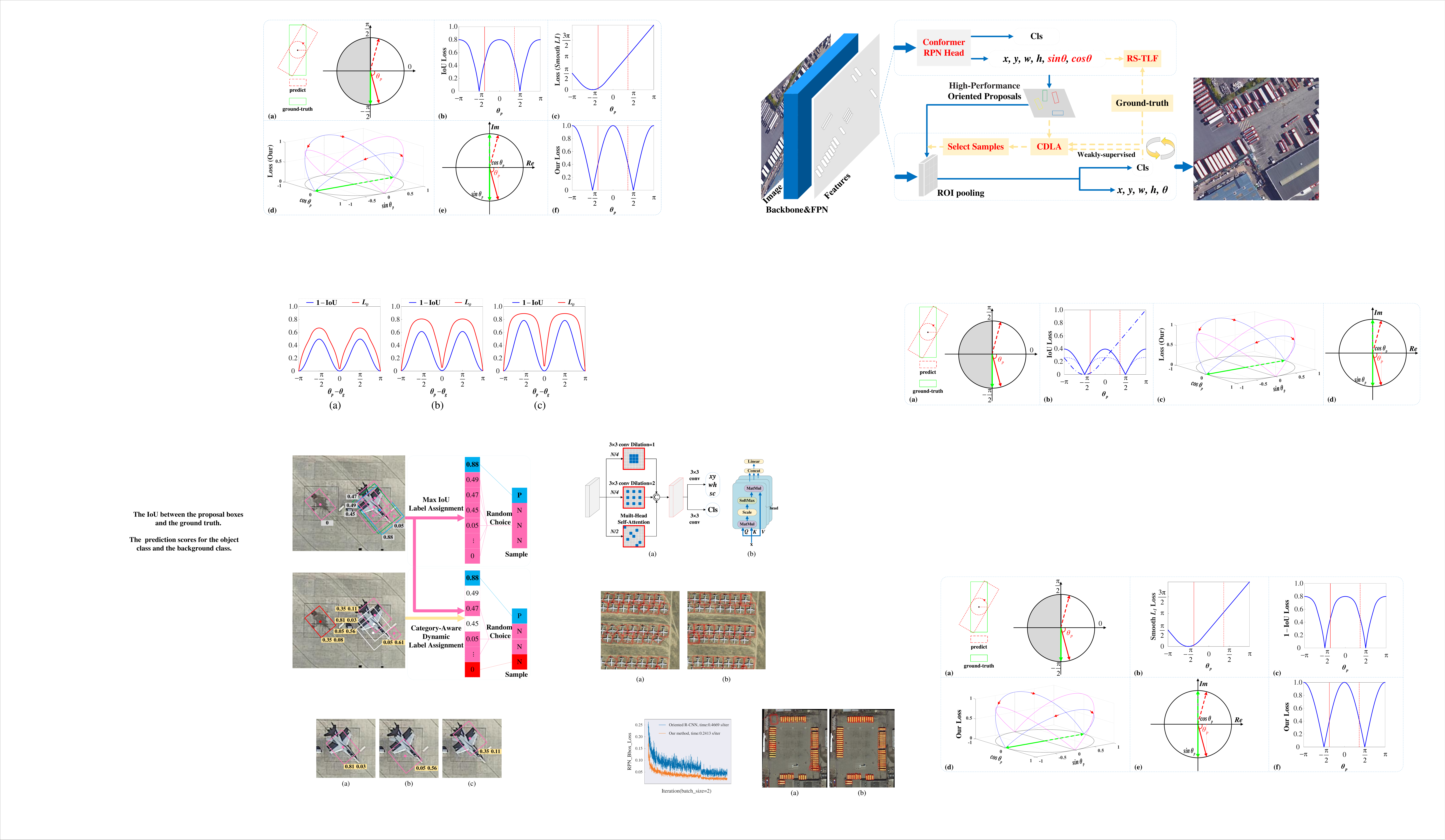}
	\caption{The horizontal coordinates indicate the angular error of the prediction and labeling. The width and height ratios of Figures (a), (b), and (c) are 2, 3, and 5, respectively.}
	\label{fig_3}
\end{figure}

The introduction only analyzes the relationship between predicted angles and IoU under fixed aspect ratios. Objects with different aspect ratios exhibit distinct sensitivities to angles and different curves of IoU versus angular deviation, as shown in Fig. \ref{fig_3}. To address this problem, an aspect ratio sensitivity factor $ \sqrt{\frac{w}{h}} $ is added to equation (\ref{eq1}). The loss in the equation (\ref{eq2}) angular branch is calculated as follows.
\begin{equation}\label{eq5}
	L{(t_{\theta}, t_{\theta}^{p})} = \sqrt{\frac{w}{h}} \left |t_{\sin \theta}^{p} t_{\cos \theta} - t_{\cos \theta}^{p} t_{\cos \theta} \right |
\end{equation}

Compared to other methods, TLF adopts complex-plane coordinates instead of angles for a consistent representation, encoding, decoding, and loss functions. The application of TLF in the model ensures high consistency between training and evaluation metrics, offering an ideal solution for angular boundary problems. Experiments show that our method is more stable during training (see Fig. \ref{fig_9}), and the results are superior to those of other methods (see Table \ref{tab:table1}). In the RetinaNet-O experiment, the regression loss for OBB replaces $ \text { smooth } L_{1} $ with $ L_{1}(x) = \left | x \right |  $ in equation (\ref{eq2}). 

\subsection{Conformer RPN Head}
\begin{figure}[!t]
	\centering
	\includegraphics[width=3.5in]{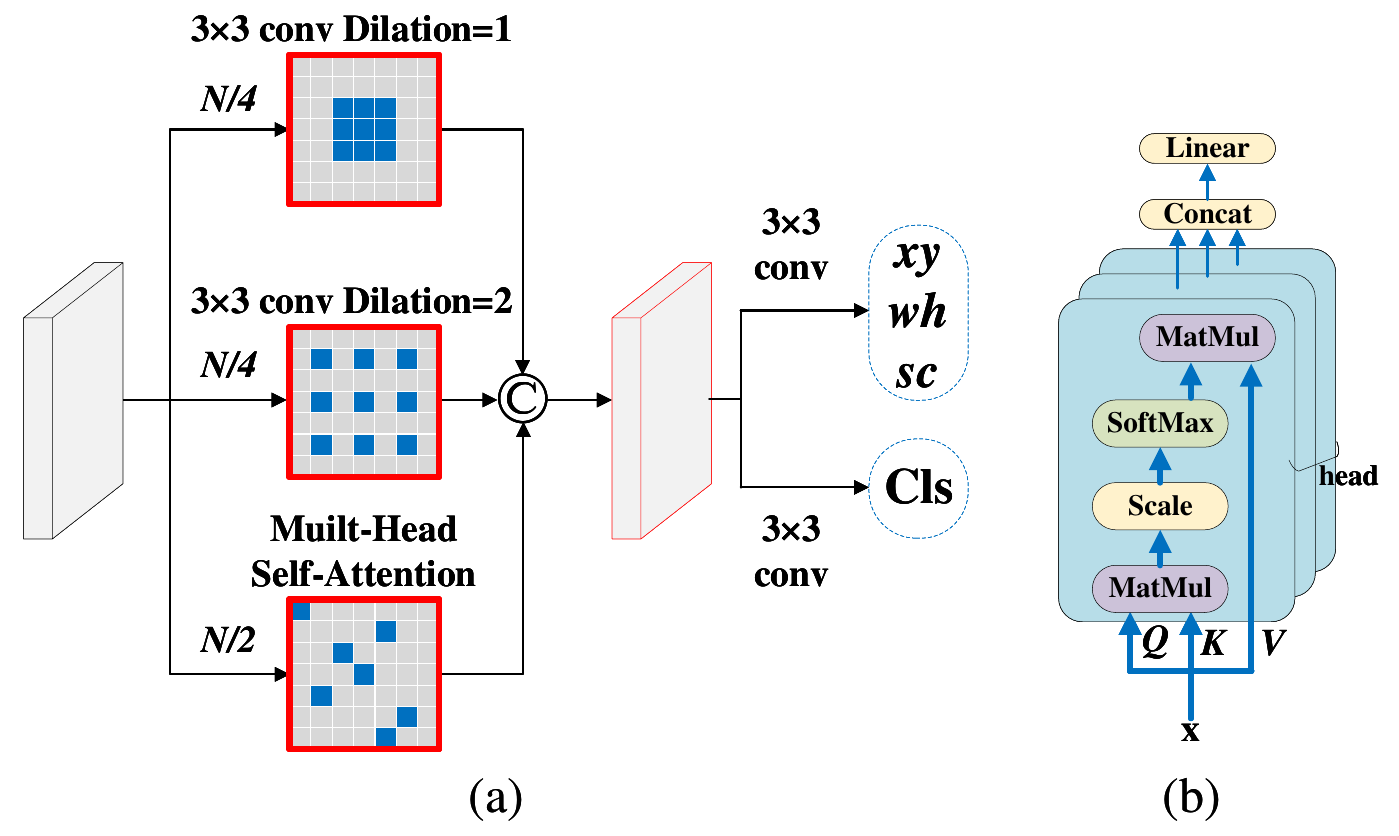}
	\caption{(a) Conformer RPN Head. (b) Multi -Head Self-Attention.}
	\label{fig_4}
\end{figure}

In a regular RPN head, the same convolution kernel is used for all channels of all input images. Due to the complex background environments and objective differences in remote sensing images, recognizing different types of objects requires varying receptive fields. To efficiently handle various ranges of detected objects, a multi-head self-attention mechanism is proposed to complement the conventional convolutional approach. A conformer RPN head containing vanilla convolution, dilation convolution, and multi-head self-attention mechanisms, as shown in Fig. \ref{fig_4}. The incorporation of multi-head self-attention enables dynamic adjustment of the receptive field for feature extraction.

Given the input feature $ X:(N, C_{in}, H, W) $. For each location $ p $ on the output of vanilla convolution and dilation convolution can be precisely described as:
\begin{equation}\label{eq6}
	\operatorname{Conv}(\boldsymbol{X})_{p}=\sum_{\boldsymbol{p}_{n} \in \mathcal{R}} \boldsymbol{x}_{(p + p_{n})} \boldsymbol{W}_{(\boldsymbol{p}_{n})}+\boldsymbol{b}
\end{equation}
where the grid $ \mathcal{R} $ defines the convolution kernel size and dilation, $ \boldsymbol{p}_{n} $ enumerates the locations in $ \mathcal{R} $.

For each location $ p $ on the output of multi-head self-attention can be precisely described as:
\begin{equation}\label{eq7}
	\operatorname{MHSA}(\boldsymbol{X})_{p}=\sum_{h \in\left[N_{h}\right]} {SA}^{(h)}\boldsymbol{(X)} \boldsymbol{W}^{(h)}+\boldsymbol{b}
\end{equation}
where $ N_{h} $ denotes the number of heads in multi-head self-attention, $ {SA} \boldsymbol{(X)} = softmax(\frac{QK^{T}}{\sqrt{d_{k} } } )V $, and the superscript $ (h) $ denotes the $ h $-th head. In practice, the $ {SA} $ and $ \boldsymbol{W}^{(h)} $ of each head are independent of each other and do not share weights, and the result is obtained by sequentially splicing and then directly cross-multiplying them. 

Since vanilla convolution and dilation volumes with fixed and horizontal receptive domains cannot align oriented objects, while multi-head self-attention can capture the global dependency and consider the information of the whole input sequence. Therefore, in this approach, 1/4 channels of features are obtained using vanilla convolution and dilation convolution each, and another 1/2 channels of features are acquired using multiple self-attention mechanisms. These features are then combined to create a new feature extraction layer that dynamically adjusts the receptive field. Considering that conformer RPN head can perceive information about various objects, they are utilized in the current detector to extract classification and regression features. Finally, classification and regression results are obtained by using vanilla convolution.

\subsection{Class-Aware Dynamic Label Assignment (CDLA)}
\begin{figure}[!t]
	\centering
	\includegraphics[width=3.5in]{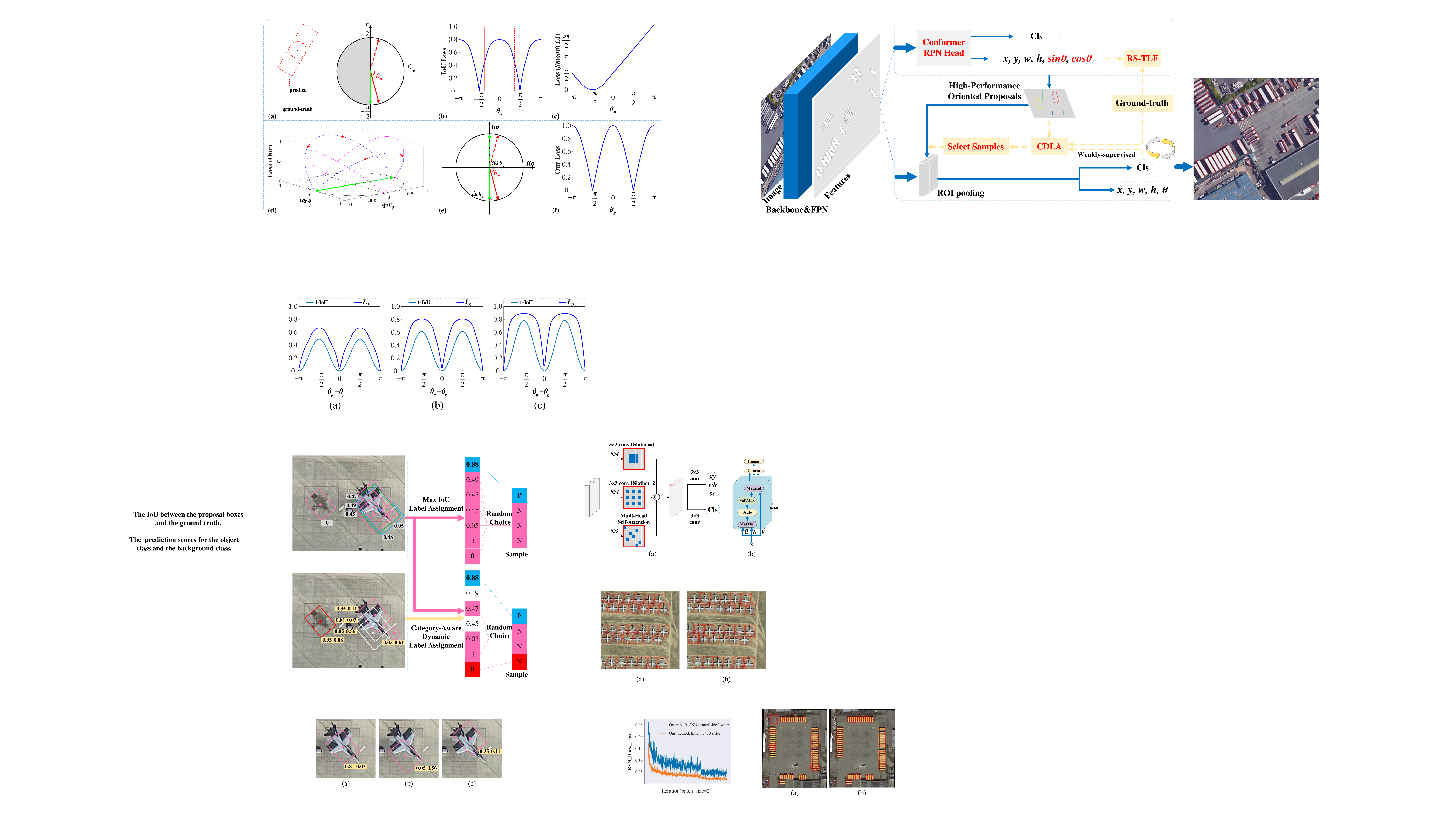}
	\caption{Illustration of the problem of label assignment for negative samples in oriented object detection. The scores in the yellow boxes represent the predicted probabilities for the current object classes and the background class in multi-object classification, denoted as $ {\mathcal{P}_{c}}_{(\textbf{TP})} $ and $ {\mathcal{P}_{c}}_{(\textbf{BK})} $ respectively. The values in the gray boxes indicate the IoU between proposal boxes and ground truth. In the maximum IoU label assignment, green, blue, and pink respectively denote ground truth, positive samples, and negative samples. In our label assignment, negative samples are divided into ignored (white), normal (pink), and focused (red) negative samples based on feedback on predicted classification values.
	}
	\label{fig_5}
\end{figure}

\begin{figure}[!t]
	\centering
	\includegraphics[width=3.5in]{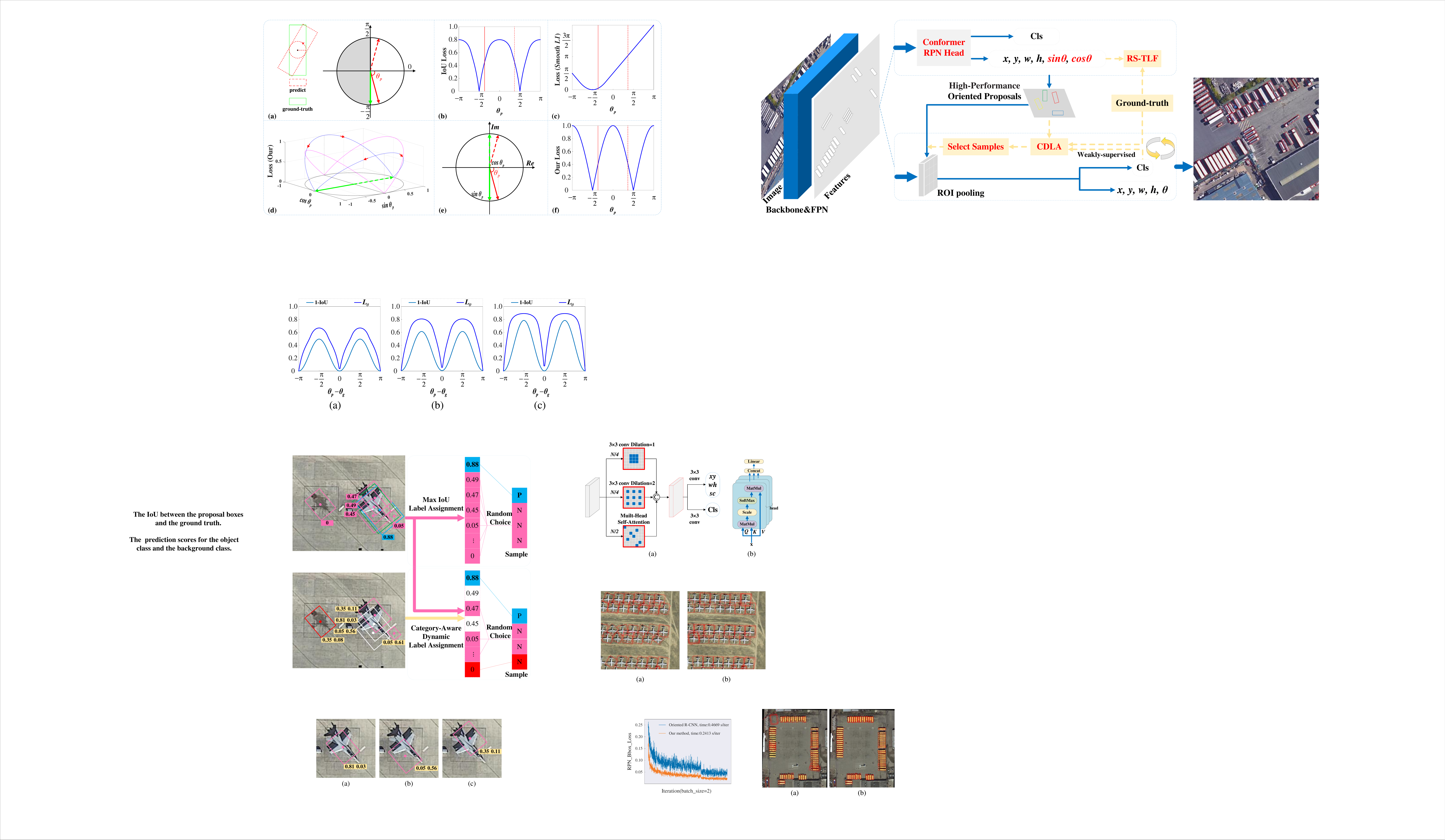}
	\caption{Category-aware weakly supervised label assignment when IoU approaches 0.5. The IoU values for images (a), (b), and (c) are 0.49, 0.45, and 0.47, respectively.
	}
	\label{fig_6}
\end{figure}

In object detection frameworks, the selection of positive and negative samples is crucial to the performance of the detector. Traditional label assignments based on IoU use a predefined IoU threshold for all objects. As illustrated by the maximum IoU label assignment in Fig. \ref{fig_5}, samples are divided into either negative or positive. However, relying solely on the IoU threshold to assign positive and negative samples is unreasonable when the IoU is between 0.4 and 0.5. As shown in Fig. \ref{fig_6}, the three oriented proposals have similar IoU values. Compared with the oriented proposal in Fig. \ref{fig_6} (a), which covers the main part of the object, Fig. 6 (b) contains more background information. Therefore, it is reasonable that the object class score in Fig. \ref{fig_6} (a) is 0.81, and the background class score in Fig. \ref{fig_6} (b) is 0.56. Compared with the IoU-based assignment that assigns the two proposals as negative samples, the proposed CDLA ignores these two proposals. Only when the object class score and the background class score are both less than 0.5, the proposed CDLA assigns the proposal as a negative sample, as shown in Fig. \ref{fig_6} (c). In summary, the proposed method achieves weakly supervised label assignment when the proposal IoU is between 0.4 and 0.5. If the classification is correct, either as the object class or background, it should be ignored during training. Otherwise, it would be assigned as a negative sample.

Conversely, the proposed method employs strong supervised label assignment when the proposal IoU is in the range of $ [0, 0.3] $ and the predicted background class score is less than 0.5. As indicated by the red boxes in Fig. \ref{fig_5}, these proposals are assigned as focus negative samples. This assignment strategy enables the detector to prioritize these negative samples during the training process.

As shown in Fig. \ref{fig_2}, the predicted categorical information by the ROI network is feedback to the ROI sampling module during training. A more representative category-aware dynamic label assignment was designed by integrating categorical information into the classical maximum IoU labeling assignment strategy. To clarify the CDLA algorithm, we summarize the process of label assignment and sample selection in \textbf{Algorithm} \ref{alg1}. Firstly, calculate the RIOU for each ground truth and proposal in the image. Secondly, the feedback category information is used to perform a detailed division of the label assignment strategy. Finally, sample selection is performed using the data obtained from the aforementioned label assignment. Through these operations, we can achieve more flexible and reliable sampling strategies to ensure consistency between classification and regression features.

\begin{algorithm}[H]
	\caption{Class-Aware Dynamic Label Assignment}\label{alg1}
	\begin{algorithmic}
		\STATE {\textbf{Input:}}
		\STATE \hspace{0.3cm}$ \mathcal{N} $ is the total number of samples
		\STATE \hspace{0.3cm}$ \mathcal{P}_{b} $ is a set of proposals via RPN outputs
		\STATE \hspace{0.3cm}$ \mathcal{G}_{b} $ is a set of ground-truth boxes on the image
		\STATE \hspace{0.3cm}$ \mathcal{P}_{c} $ is a set of classification score matrix via ROI outputs
		\STATE {\textbf{Output:}}
		\STATE \hspace{0.3cm}$ \mathcal{PS} $ is a set of positive samples
		\STATE \hspace{0.3cm}$ \mathcal{NS} $ is a set of negative samples, where $ \mathcal{NS}_{f},\mathcal{NS}_{n} $ represent focus and normal negative samples.		
		\STATE {\textbf{Class-Aware Dynamic Label Assignment}}
		\STATE \hspace{0.3cm}$ \mathcal{A} $ is a set of all label assignment
		\STATE 1:\hspace{0.0cm} \textbf{for} each ground-truth $ g \in \mathcal{G}_{b} $ \textbf{do}
		\STATE 2:\hspace{0.5cm} compute IoU between $ \mathcal{P}_{b} $ and $ \mathcal{G}_{b} $: $ \mathcal{I}=IoU( \mathcal{P}_{b}, \mathcal{G}_{b}) $;
		\STATE 3:\hspace{0.5cm} \textbf{if} $ (\mathcal{I} \in [0.5,1.0]) $ \textbf{then}
		\STATE 4:\hspace{1.0cm} $ \mathcal{A}(\mathcal{I}) \in \mathcal{PS} $;
		\STATE 5:\hspace{0.5cm} \textbf{else if} $ (\mathcal{I} \in [0.4,0.5) $ \textbf{then}
		\STATE 6:\hspace{1.0cm} \textbf{if} $ ({\mathcal{P}_{c}}_{(\textbf{TP and BK}})<0.5) $ \textbf{then}
		\STATE 7:\hspace{1.5cm} $ \mathcal{A}(\mathcal{I}) \in \mathcal{NS}_{n} $;
		\STATE 8:\hspace{0.5cm} \textbf{else if} $ ((\mathcal{I} \in [0,0.3] ) \textbf{and} ({\mathcal{P}_{c}}_{(\textbf{BK})}<0.5)) $ \textbf{then}
		\STATE 9:\hspace{1.0cm} $ \mathcal{A}(\mathcal{I}) \in \mathcal{NS}_{f} $;
		\STATE \hspace{-0.15cm}10:\hspace{0.5cm} \textbf{else} 
		\STATE \hspace{-0.15cm}11:\hspace{1.0cm} $ \mathcal{A}(\mathcal{I}) \in \mathcal{NS}_{n} $; 
		\STATE {\textbf{Sample}}
		\STATE 1:\hspace{0.0cm} $ \mathcal{PS} = random \hspace{0.1cm} choice(\mathcal{PS}, \textbf{min}(\textit{\textbf{len}}(\mathcal{PS}),\mathcal{N}/4)) $ 
		\STATE 2:\hspace{0.0cm} $ \mathcal{NS}_{f} = random \hspace{0.1cm} choice(\mathcal{NS}_{f}, \textbf{min}(\textit{\textbf{len}}(\mathcal{NS}_{f}),\mathcal{N}/8)) $ 
		\STATE 3:\hspace{0.0cm} $ L = \mathcal{N} - \textit{\textbf{len}}(\mathcal{PS}) - \textit{\textbf{len}}(\mathcal{NS}_{f}) $
		\STATE 4:\hspace{-0.0cm} $ \mathcal{NS}_{n} = random \hspace{0.1cm} choice(\mathcal{NS}_{n}, \textbf{min}(\textit{\textbf{len}}(\mathcal{NS}_{n}),L)) $ 	
		\STATE 5:\hspace{-0.0cm} $ \mathcal{NS} = \mathcal{NS}_{f} \cup \mathcal{NS}_{n} $ 		
		\STATE 6:\hspace{0.0cm} \textbf{return}  $ \mathcal{PS}, \mathcal{NS} $
	\end{algorithmic}
\end{algorithm}

\section{EXPERIMENTS AND RESULTS}
\subsection{Datasets}
We selected four public datasets for evaluating oriented object detection, namely DOTA-v1.0\cite{CVPR2018_DOTA}, DOTA-v1.5\cite{CVPR2018_DOTA}, DIOR-R \cite{P&RS2020_DIOR, TGRS2022_AOPG}, and HRSC2016\cite{ICPRAM2017_HRSC}. Details are as follows.

\subsubsection{DOTA}
The DOTA-v1.0 dataset is a comprehensive dataset designed for oriented object detection in aerial images. It comprises 2806 images captured by different sensors and platforms, with image sizes varying from $ 800 \times 800 $ to $ 4000 \times 4000 $ pixels. These images are divided into $ 1024 \times 1024 $ sub-images with an overlap of 200 pixels. The fully annotated DOTA-v1.0 benchmark contains 188 282 instances across 15 common object categories, each labeled with an arbitrary quadrilateral. All reported results for DOTA were obtained by evaluating the test set on the official evaluation server.

Compared to DOTA-v1.0, DOTA-v1.5 introduces a new category called Container Crane (CC) and increases the number of micro-instances with a size smaller than 10 pixels. It contains a total of 403 318 instances. 

\subsubsection{DIOR-R}
DIOR-R consists of 23 463 images and 192 518 instances, covering 20 object classes. The dataset contains fixed-size images of $ 800 \times 800 $ pixels, featuring a large variation in object sizes (ranging from 0.5 to 30 meters) in terms of spatial resolution, as well as significant inter-class and intra-class size variability. The training and validation sets contain a total of 11 725 images and 68 073 instances, while the testing set comprises 11 738 images and 124 445 instances.

\subsubsection{HRSC2016}
HRSC2016 is a dataset for arbitrary-oriented ship detection in aerial images, containing 1061 images from six famous harbors. The image sizes range from $ 300 \times 300 $ to $ 1500 \times 900 $ pixels, and we resize them to $ 800 \times 800 $ pixels. The combined training set (436 images) and validation set (181 images) are used for training, while the remaining 455 images are used for testing. For detection accuracy on HRSC2016, we adopt the mAP as the evaluation criterion, consistent with PASCAL VOC 2007 and VOC 2012.

\subsection{Implementation Details}
All experiments were implemented using a single NVIDIA RTX 3090 with a batch size of 2, trained with half-precision on the MMRotate toolbox \cite{ACMMM2022_MMRotate}. The training schedule uses ``$ 1 \times $'' for the DOTA and DIOR-R datasets, and ``$ 3 \times $'' for the HRSC2016 dataset. Competitive results have already been achieved using ResNet as the backbone network. Further validation was performed with ConvNeXt \cite{CVPR2022_ConvNeXt} to demonstrate the generalization and scalability of the framework. When using ResNet, the entire network was optimized using the SGD algorithm with a learning rate of 0.005, momentum of 0.9, weight decay of 0.0001, and a learning rate warm-up method with 500 iterations. When using ConvNeXt-T, the network was optimized using the AdamW algorithm with a learning rate of 0.0001, weight decay of 0.05, and a learning rate warm-up method with 1000 iterations. 

\begin{table*}[!ht]
	\caption{\textsc{Comparison with the SOTA methods using ResNet as the backbone on the DOTA-v1.0 dataset. $ \ast $ means our reimplementation based on the mmdrotate. $ \ddag $ means our multiscale traning and testing \label{tab:table1}}}
	\renewcommand\arraystretch{1.25}
	\centering
	\resizebox{\linewidth}{!}{ 
		\begin{tabular}{c|c|c c c c c c c c c c c c c c c c}
			\hline
			Methods & Backbone & PL & BD & BR & GTF & SV & LV & SH & TC & BC & ST & SBF & RA & HA & SP & HC & mAP \\ 
			\hline
			RetinaNet-O \cite{ICCV2017_RetinaNet} & R-50 & 88.67  & 77.62  & 41.81  & 58.17  & 74.58  & 71.64  & 79.11  & 90.29  & 82.18  & 74.32  & 54.75  & 60.60  & 62.57  & 69.67  & 60.64  & 68.43  \\ 
			CSL-RetinaNet-O \cite{ECCV2020_CSL} & R-50 & 89.33  & 79.67  & 40.83  & 69.95  & 77.71  & 62.08  & 77.46  & 90.87  & 82.87  & 82.03  & 60.07  & 65.27  & 53.58  & 64.03  & 46.62  & 69.49  \\ 
			KLD-RetinaNet-O \cite{NeurIPS2021_KLD} & R-50 & 89.50  & 79.91  & 39.92  & 70.40  & 78.04  & 64.24  & 82.79  & 90.90  & 81.80  & 83.02  & 57.63  & 63.52  & 56.63  & 65.13  & 50.04  & 70.23  \\ 
			PSCD-RetinaNet-O \cite{CVPR2023_PSC} & R-50 & 89.32 & 82.29 & 37.92 & 71.52 & 78.4 & 66.33 & 78.01 & 90.89 & 84.21 & 80.63 & 60.22 & 64.73 & 59.69 & 68.37 & 53.85 & 71.09 \\ 
			TLF-RetinaNet-O(Our) & R-50 & 89.34 & 82.30 & 39.60 & 71.09 & 79.07 & 66.76 & 78.13 & 90.86 & 83.89 & 81.03 & 58.49 & 65.51 & 59.30 & 70.35 & 52.17 & \textbf{71.19} \\
			\hline
			FR-O$ ^{\ast} $ \cite{TPAMI2017_FasterR-CNNa} & R-50 & 89.25  & 82.40  & 50.02  & 69.37  & 78.17  & 73.56  & 85.92  & 90.90  & 84.08  & 85.49  & 57.58  & 60.98  & 66.25  & 69.23  & 57.74  & 73.40  \\ 
			RoI Trans$ ^{\ast} $ \cite{CVPR2019_RoITrans} & R-50 & 88.65  & 82.60  & 52.53  & 70.87  & 77.93  & 76.67  & 86.87  & 90.71  & 83.83  & 82.51  & 53.95  & 67.61  & 74.67  & 68.75  & 61.03  & 74.61  \\ 
			AOPG \cite{TGRS2022_AOPG} & R-50 & 89.27  & 83.49  & 52.50  & 69.97  & 73.51  & 82.31  & 87.95  & 90.89  & 87.64  & 84.71  & 60.01  & 66.12  & 74.19  & 68.30  & 57.85  & 75.24  \\ 
			DODet \cite{TGRS2022_DODet} & R-50 & 89.34  & 84.31  & 51.39  & 71.04  & 79.04  & 82.86  & 88.15  & 90.90  & 86.88  & 84.91  & 62.69  & 67.63  & 75.47  & 72.22  & 45.54  & 75.49  \\ 
			Oriented R-CNN \cite{ICCV2021_ORCNN} & R-50 & 89.46  & 82.12  & 54.78  & 70.86  & 78.93  & 83.00  & 88.20  & 90.90  & 87.50  & 84.68  & 63.97  & 67.69  & 74.94  & 68.84  & 52.28  & 75.87  \\ 
			QPDet \cite{TGRS2023_QPDet} & R-50 & 89.55  & 83.66  & 54.06  & 73.93  & 78.93  & 83.08  & 88.29  & 90.89  & 86.60  & 84.80  & 62.03  & 65.55  & 74.16  & 70.09  & 58.16  & 76.25  \\ 
			Our  & R-50 & 89.54  & 83.14  & 55.32  & 71.56  & 80.09  & 83.58  & 88.20  & 90.90  & 87.93  & 85.77  & 65.69  & 66.30  & 74.80  & 71.29  & 63.72  & \textbf{77.19} \\ 
			\hline
			Gliding Vertex $ \ddag $ \cite{TPAMI2021_GlidingVertex} & R-101 & 89.64  & 85.00  & 52.26  & 77.34  & 73.01  & 73.14  & 86.82  & 90.74  & 79.02  & 86.81  & 59.55  & 70.91  & 72.94  & 70.86  & 57.32  & 75.02  \\
			SCRDet++ $ \ddag $ \cite{TPAMI2023_SCRDetplusplus} & R-101 & 90.05  & 84.39  & 55.44  & 73.99  & 77.54  & 71.11  & 86.05  & 90.67  & 87.32  & 87.08  & 69.62  & 68.90  & 73.74  & 71.29  & 65.08  & 76.81  \\  
			Oriented R-CNN $ \ddag $ \cite{ICCV2021_ORCNN} & R-50 & 89.84  & 85.43  & 61.09  & 79.82  & 79.71  & 85.35  & 88.82  & 90.88  & 86.68  & 87.73  & 72.21  & 70.08  & 82.42  & 78.18  & 74.11  & 80.87  \\ 
			QPDet $ \ddag $ \cite{TGRS2023_QPDet} & R-50 & 90.14  & 85.31  & 60.98  & 79.92  & 80.21  & 85.04  & 88.80  & 90.87  & 86.45  & 88.04  & 70.88  & 71.72  & 82.99  & 80.55  & 73.19  & 81.00  \\ 
			Our $ \ddag $  & R-50 & 89.69 & 85.25 & 60.37 & 81.78 & 80.47 & 85.65 & 88.80 & 90.87 & 86.45 & 87.75 & 72.22 & 72.43 & 78.78 & 81.17 & 75.01 & \textbf{81.11} \\ 
			\hline
		\end{tabular}
	}
\end{table*}

\begin{table*}[!ht]
	\caption{\textsc{Comparison with SOTA methods on the DOTA-v1.0 dataset. $ \ddag $ means our multiscale traning and testing \label{tab:table2}}}
	\renewcommand\arraystretch{1.25}
	\centering
	\resizebox{\linewidth}{!}{ 
		\begin{tabular}{c|c|c c c c c c c c c c c c c c c c}
			\hline
			Methods & Backbone & PL & BD & BR & GTF & SV & LV & SH & TC & BC & ST & SBF & RA & HA & SP & HC & mAP \\ 
			\hline
			ReDet\cite{CVPR2021_ReDet} & ReR50-ReFPN & 88.79  & 82.64  & 53.97  & 74.00  & 78.13  & 84.06  & 88.04  & 90.89  & 87.78  & 85.75  & 61.76  & 60.39  & 75.96  & 68.07  & 63.59  & 76.25  \\
			Oriented R-CNN \cite{ICCV2021_ORCNN} & ARC-R50 \cite{ICCV2023_ARC} & 89.40  & 82.48  & 55.33  & 73.88  & 79.37  & 84.05  & 88.06  & 90.90  & 86.44  & 84.83  & 63.63  & 70.32  & 74.29  & 71.91  & 65.43  & 77.35  \\ 
			Oriented R-CNN \cite{ICCV2021_ORCNN} & PKINet-S \cite{CVPR2024_PKINet} & 89.72  & 84.20  & 55.81  & 77.63  & 80.25  & 84.45  & 88.12  & 90.88  & 87.57  & 86.07  & 66.86  & 70.23  & 77.47  & 73.62  & 62.94  & 78.39  \\ 
			Oriented R-CNN \cite{ICCV2021_ORCNN} $ \ddag $ & LSKNet-S\cite{ICCV2023_LSKNet} & 89.57  & 86.34  & 63.13  & 83.67  & 82.20  & 86.10  & 88.66  & 90.89  & 88.41  & 87.42  & 71.72  & 69.58  & 78.88  & 81.77  & 76.52  & 81.64  \\ 
			Our  & ConvNeXt-T\cite{CVPR2022_ConvNeXt} & 89.60 & 85.83 & 56.19 & 77.10 & 80.25 & 84.98 & 88.40 & 90.85 & 87.99 & 86.14 & 69.80 & 70.32 & 76.97 & 74.09 & 65.49 & 78.93 \\ 
			Our $ \ddag $  & ConvNeXt-T\cite{CVPR2022_ConvNeXt} & 89.59 & 86.93 & 61.91 & 82.94 & 80.37 & 85.90 & 88.62 & 90.85 & 87.23 & 87.83 & 72.54 & 73.92 & 79.35 & 80.40 & 81.89 & \textbf{82.02} \\ 
			\hline
		\end{tabular}
	}
\end{table*}

\subsection{Comparisons With State-of-the-Art Methods}
\subsubsection{Results on DOTA}
\begin{table*}[!ht]
	\caption{\textsc{Main results on DOTA-v1.5 test set under single-scale training and testing. \label{tab:table3}}}
	\renewcommand\arraystretch{2.0}
	\centering
	\resizebox{\linewidth}{!}{ 
		\begin{tabular}{c c c c c c c c c c c c c}
			\hline
			Method & RetinaNet-O\cite{ICCV2017_RetinaNet} & FR-O\cite{TPAMI2017_FasterR-CNNa} & RoI Trans\cite{CVPR2019_RoITrans} & AOPG\cite{TGRS2022_AOPG} & GGHL\cite{TIP2022_GGHL} & Oriented Rep\cite{CVPR2022_OrientedRepPoints} & DCFL\cite{CVPR2023_DCFL} & our(R-50) & ReDet\cite{CVPR2021_ReDet} & LSKNet-S\cite{ICCV2023_LSKNet} & PKINet-S\cite{CVPR2024_PKINet} & our(ConvNeXt-T\cite{CVPR2022_ConvNeXt}) \\ \hline
			mAP & 59.16 & 62.00 & 63.87 & 64.41 & 66.48 & 66.71 & 66.80 & 67.82 & 66.86 & 70.26 & 71.47 & \textbf{71.99} \\ 
			\hline
		\end{tabular}
	}
\end{table*}

\begin{figure*}[!t]
	\centering
	\includegraphics[width=6.5in]{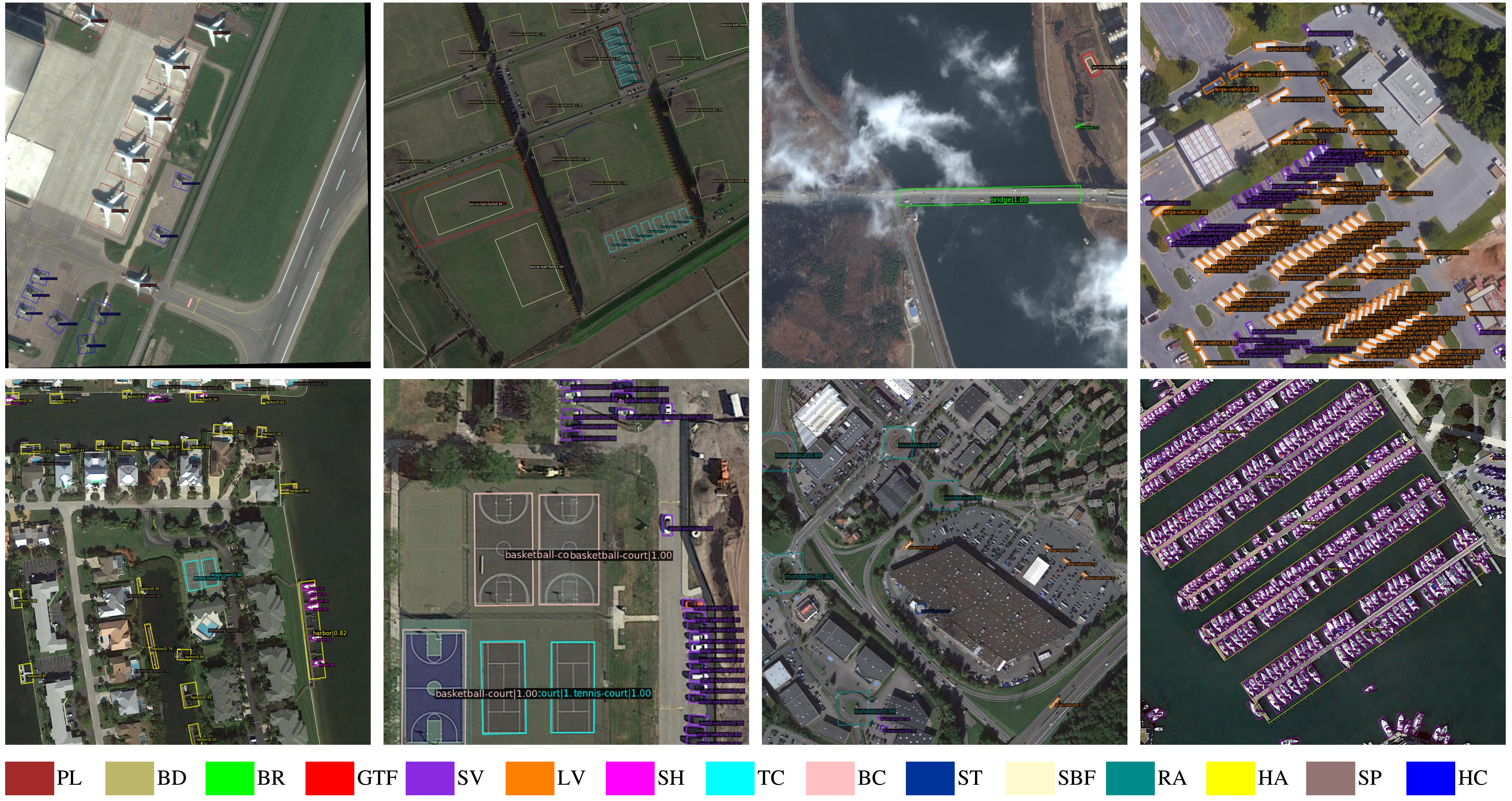}
	\caption{Detection results of our method on the DOTA dataset.}
	\label{fig_7}
\end{figure*}

Table \ref{tab:table1} presents a comparison of our method with the SOTA methods using ResNet as the backbone on the DOTA-v1.0 dataset. RetinaNet-O with ResNet-50 as the backbone, using only TLF, achieves 71.19\% mAP, surpassing many other outstanding loss functions. Using ResNet-50 as the backbone network, our method achieved 77.19\% and 81.11\% mAP for single-scale and multi-scale, respectively. In the single-scale setting, it surpassed the advanced Oriented R-CNN and ReDet with the mAP gains of 1.32\% and 0.94\%, respectively. Table \ref{tab:table2} presents a comparison of our method using ConvNeXt-T as the backbone with other SOTA methods on the DOTA-v1.0 dataset. Our method achieves a SOTA result of 79.19\% mAP for single-scale. In the multiscale setting, the mAP reached 82.02\%, surpassing the SOTA results achieved with CNN as the backbone. We visualized some qualitative detection results, as shown in Fig. \ref{fig_7}.

Additionally, comparative experiments were conducted on the new version of the DOTA-v1.5 dataset. Table \ref{tab:table3} lists the quantitative results of DOTA-v1.5. In the single-scale setting, our proposed method achieves 67.82\% and 71.99\% mAP using ResNet-50 and ConvNeXt-T as the backbone networks, respectively. Faced with challenging new situations, our method demonstrated satisfactory results, proving its robustness.

\subsubsection{Results on the DIOR-R}
\begin{figure*}[!t]
	\centering
	\includegraphics[width=7.0in]{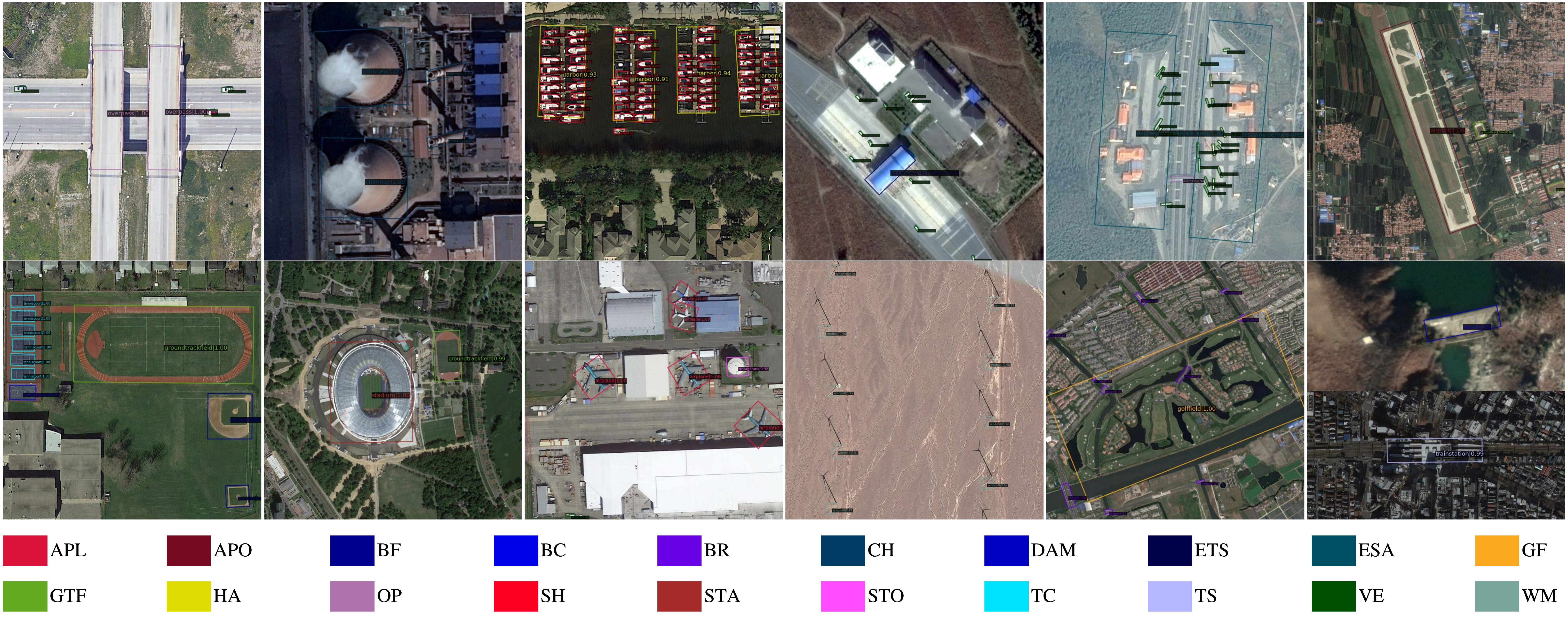}
	\caption{Detection results of our method on the DIOR-R dataset.}
	\label{fig_8}
\end{figure*}

DIOR-R is a dataset annotated with five-parameter, and the results from this dataset are more indicative of the effectiveness of our method. The results of eight oriented detectors are reported in \ref{tab:table6}. Our proposed method achieves 65.93\% and 69.87\% mAP using ResNet-50 and ConvNeXt-T as the backbone networks, respectively. From the evaluation metrics and visualization results, it can be observed that our method achieved good performance, particularly on objects with large aspect ratios. Compared to other directed object detectors, our method shows strong competitiveness. Fig. \ref{fig_8} visualizes some detection results from the DIOR-R dataset.

\begin{table*}[!ht]
	\caption{\textsc{Comparison with the SOTA methods on the DIOR-R dataset\label{tab:table6}}}
	\renewcommand\arraystretch{1.5}
	\centering
	\resizebox{\linewidth}{!}{ 
		\begin{tabular}{c|c|c c c c c c c c c c c c c c c c c c c c c}
			\hline
			Methods & Backbone & APL & APO & BF & BC & BR & CH & DAM & ETS & ESA & GF & GTF & HA & OP & SH & STA & STO & TC & TS & VE & WM & mAP \\ 
			\hline
			RetinaNet-O\cite{ICCV2017_RetinaNet} & R-50 & 61.49  & 28.52  & 73.57  & 81.17  & 23.98  & 72.54  & 19.94  & 72.39  & 58.20  & 69.25  & 79.54  & 32.14  & 44.87  & 77.71  & 67.57  & 61.09  & 81.46  & 47.33  & 38.01  & 60.24  & 57.55  \\ 
			FR-O\cite{TPAMI2017_FasterR-CNNa} & R-50 & 62.79  & 26.80  & 71.72  & 80.91  & 34.20  & 72.57  & 18.95  & 66.45  & 65.75  & 66.63  & 79.24  & 34.95  & 48.79  & 81.14  & 64.34  & 71.21  & 81.44  & 47.31  & 50.46  & 65.21  & 59.54  \\ 
			Gliding Vertex\cite{TPAMI2021_GlidingVertex} & R-50 & 63.35  & 28.87  & 74.96  & 81.33  & 33.88  & 74.31  & 19.58  & 70.72  & 64.70  & 72.30  & 78.68  & 37.22  & 49.64  & 80.22  & 69.26  & 61.13  & 81.49  & 44.76  & 47.71  & 65.04  & 60.06  \\ 
			RoI Trans\cite{CVPR2019_RoITrans} & R-50 & 63.34  & 37.88  & 71,78 & 87.53  & 40.68  & 72.60  & 26.86  & 78.71  & 68.09  & 68.96  & 82.74  & 47.71  & 55.61  & 81.21  & 78.23  & 70.26  & 81.61  & 54.86  & 43.27  & 65.52  & 63.87  \\ 
			QPDet\cite{TGRS2023_QPDet} & R-50 & 63.22  & 41.39  & 71.97  & 88.55  & 41.23  & 72.63  & 28.82  & 78.90  & 69.00  & 70.07  & 83.01  & 47.83  & 55.54  & 81.23  & 72.15  & 62.66  & 89.05  & 58.09  & 43.38  & 65.36  & 64.20  \\ 
			AOPG\cite{TGRS2022_AOPG} & R-50 & 62.39  & 37.79  & 71.62  & 87.63  & 40.90  & 72.47  & 31.08  & 65.42  & 77.99  & 73.20  & 81.94  & 42.32  & 54.45  & 81.17  & 72.69  & 71.31  & 81.49  & 60.04  & 52.38  & 69.99  & 64.41  \\ 
			DODet\cite{TGRS2022_DODet} & R-50 & 63.40  & 43.35  & 72.11  & 81.32  & 43.12  & 72.59  & 33.32  & 78.77  & 70.84  & 74.15  & 75.47  & 48.00  & 59.31  & 85.41  & 74.04  & 71.56  & 81.52  & 55.47  & 51.86  & 66.40  & 65.10 \\ 
			\hline
			Our & R-50 & 71.74  & 40.87  & 79.29  & 89.65  & 42.94  & 72.63  & 34.50  & 68.72  & 79.58  & 71.78  & 83.03  & 41.92  & 57.64  & 81.29  & 79.87  & 62.62  & 89.45  & 56.61  & 48.83  & 65.57  & 65.93 \\ 
			Our & ConvNeXt-T\cite{CVPR2022_ConvNeXt} & 72.19  & 52.12  & 80.44  & 90.11  & 47.91  & 80.56  & 36.02  & 73.87  & 88.11  & 79.29  & 83.84  & 45.96  & 62.21  & 81.27  & 82.87  & 70.04  & 89.51  & 64.66  & 50.14  & 66.40  & \textbf{69.87} \\ 
			\hline
		\end{tabular}
	}
\end{table*}

\subsubsection{Results on HRSC2016}
HRSC2016 contains numerous ship images with large aspect ratios in arbitrary directions, and the angle's impact on IoUs is significant. The experimental results using RetinaNet-O as the baseline are presented in Table \ref{tab:table4}. The experimental results for SOTA are presented in Table \ref{tab:table5}. Our method not only achieved improvements in mAP50 but also demonstrated significantly higher accuracy in mAP75 and mAP50:95 compared to similar methods. Specifically, using only our TLF on RetinaNet-O resulted in an mAP50 of 86.70\%, mAP75 of 72.90\%, and mAP50:95 of 61.34\%. In the two-stage approach, our method achieves results of 90.70\% and 90.89\% on VOC2007, and 98.02\% and 98.77\% on VOC2012, respectively, using R-50 and ConvNeXt-T\cite{CVPR2022_ConvNeXt} backbone networks. Our method achieves SOTA results.
\begin{table}[!t]
	\caption{\textsc{Results of the SOTA method on HRSC2016 using RetinaNet-O as a baseline. \label{tab:table4}}}
	\renewcommand\arraystretch{1.25}
	\centering
	\resizebox{\linewidth}{!}{ 
		\begin{tabular}{c c c c c c}
			\hline
			Method & Backbone & mAP50 & mAP75 & mAP50:95 \\ 
			\hline
			RetinaNet-O\cite{ICCV2017_RetinaNet} & R-50 & 84.80 & 58.10 & 52.06 \\ 
			RetinaNet-O\cite{ICCV2017_RetinaNet} & ARC-R50\cite{ICCV2023_ARC} & 85.10 & 60.20 & 53.97 \\ 
			KLD-RetinaNet-O\cite{NeurIPS2021_KLD} & R-50 & 85.85 & 58.76 & 53.40 \\ 
			CSL-RetinaNet-O\cite{ECCV2020_CSL} & R-50 & 84.87 & 38.75 & 44.17 \\
			PSC-RetinaNet-O\cite{CVPR2023_PSC} & R-50 & 85.65 & 61.30 & 54.14 \\ 
			PSCD-RetinaNet-O\cite{CVPR2023_PSC} & R-50 & 85.53 & 59.57 & 53.20 \\ 
			TLF-RetinaNet-O(Our) & R-50 & \textbf{86.70} & \textbf{72.90} & \textbf{61.34} \\ 
			\hline
		\end{tabular}
	}
\end{table}

\begin{table}[!t]
	\caption{\textsc{Results of the SOTA method on HRSC2016 \label{tab:table5}}}
	\renewcommand\arraystretch{1.25}
	\centering
	\resizebox{\linewidth}{!}{ 
		\begin{tabular}{c c c c c}
			\hline
			Method & Backbone  & mAP50 (VOC 07) & mAP50 (VOC 12) \\ 
			FR-O\cite{TPAMI2017_FasterR-CNNa} & R-50 & 87.20 & 89.51 \\
			DODet\cite{TGRS2022_DODet} & R-101 & \textbf{90.89} & 97.14 \\
			QPDet\cite{TGRS2023_QPDet} & R-50 & 90.47 & 96.60 \\
			AOPG\cite{TGRS2022_AOPG} & R-50 & 90.34 & 96.22 \\
			ReDet\cite{CVPR2021_ReDet} & ReR50-ReFPN & 90.46 & 97.63 \\
			Oriented R-CNN\cite{ICCV2021_ORCNN} & R-50 & 90.40 & 96.50 \\
			Oriented R-CNN\cite{ICCV2021_ORCNN} & LSKNet-S\cite{ICCV2023_LSKNet} & 90.65 & 98.46 \\
			Oriented R-CNN\cite{ICCV2021_ORCNN} & PKINet-S\cite{CVPR2024_PKINet} & 90.70 & 98.54 \\
			Our & R-50 & 90.70 & 98.02 \\
			Our & ConvNeXt-T\cite{CVPR2022_ConvNeXt} & \textbf{90.89} & \textbf{98.77} \\
			\hline
		\end{tabular}
	}
\end{table}

\subsection{Ablation Studies}
\begin{table*}[!ht]
	\caption{\textsc{Ablation study of proposed modules on DOTA-v1.0 dataset test set. Baseline means the Faster R-CNN-O, i.e., RPN uses HBB and ROI uses OBB. N-Baseline means that the OBB is used in both RPN and ROI\label{tab:table7}}}
	\renewcommand\arraystretch{1.25}
	\centering
	\resizebox{\linewidth}{!}{ 
		\begin{tabular}{c c c c c c c c c c c c c c c c c c c c}
			\hline
			Methods & TLF & \makecell[c]{Conformer \\ RPN Head} & CDLA & PL & BD & BR & GTF & SV & LV & SH & TC & BC & ST & SBF & RA & HA & SP & HC & mAP \\ 
			\hline
			Baseline & - & - & - & 89.25  & 82.40  & 50.02  & 69.37  & 78.17  & 73.56  & 85.92  & 90.90  & 84.08  & 85.49  & 57.58  & 60.98  & 66.25  & 69.23  & 57.74  & 73.40  \\ 
			N-Baseline & - & - & - & 89.24  & 83.09  & 51.22  & 70.58  & 78.39  & 82.60  & 88.18  & 90.90  & 85.06  & 84.83  & 58.86  & 61.57  & 68.06  & 67.61  & 55.11  & 74.35  \\ 
			\hline
			\multirow{5}*{\makecell[c]{Proposed \\ Method}} & $ \checkmark $ & - & - & 89.27  & 83.38  & 53.05  & 72.45  & 78.98  & 82.62  & 87.90  & 90.89  & 85.79  & 84.87  & 65.98  & 62.56  & 73.78  & 70.41  & 60.72  & 76.18  \\ 
			~  & - & - & $ \checkmark $ & 89.38  & 83.81  & 53.37  & 72.87  & 79.74  & 82.07  & 87.96  & 90.90  & 86.44  & 85.95  & 63.72  & 65.29  & 74.45  & 70.93  & 58.31  & 76.34  \\ 
			~  & $ \checkmark $ & $ \checkmark $ & - & 89.32  & 80.67  & 53.60  & 71.91  & 79.03  & 82.60  & 88.09  & 90.88  & 86.97  & 85.12  & 67.27  & 67.23  & 74.37  & 70.02  & 63.13  & 76.68  \\ 
			~  & $ \checkmark $ & - & $ \checkmark $ & 89.32  & 83.50  & 53.31  & 70.99  & 79.90  & 82.46  & 88.27  & 90.90  & 87.26  & 85.44  & 63.14  & 67.40  & 74.89  & 72.13  & 60.60  & 76.63  \\ 
			~  & $ \checkmark $ & $ \checkmark $ & $ \checkmark $ & 89.54  & 83.14  & 55.32  & 71.56  & 80.09  & 83.58  & 88.20  & 90.90  & 87.93  & 85.77  & 65.69  & 66.30  & 74.80  & 71.29  & 63.72  & 77.19  \\ 
			\hline
		\end{tabular}
	}
\end{table*}

\begin{figure}[!t]
	\centering
	\includegraphics[width=3.0in]{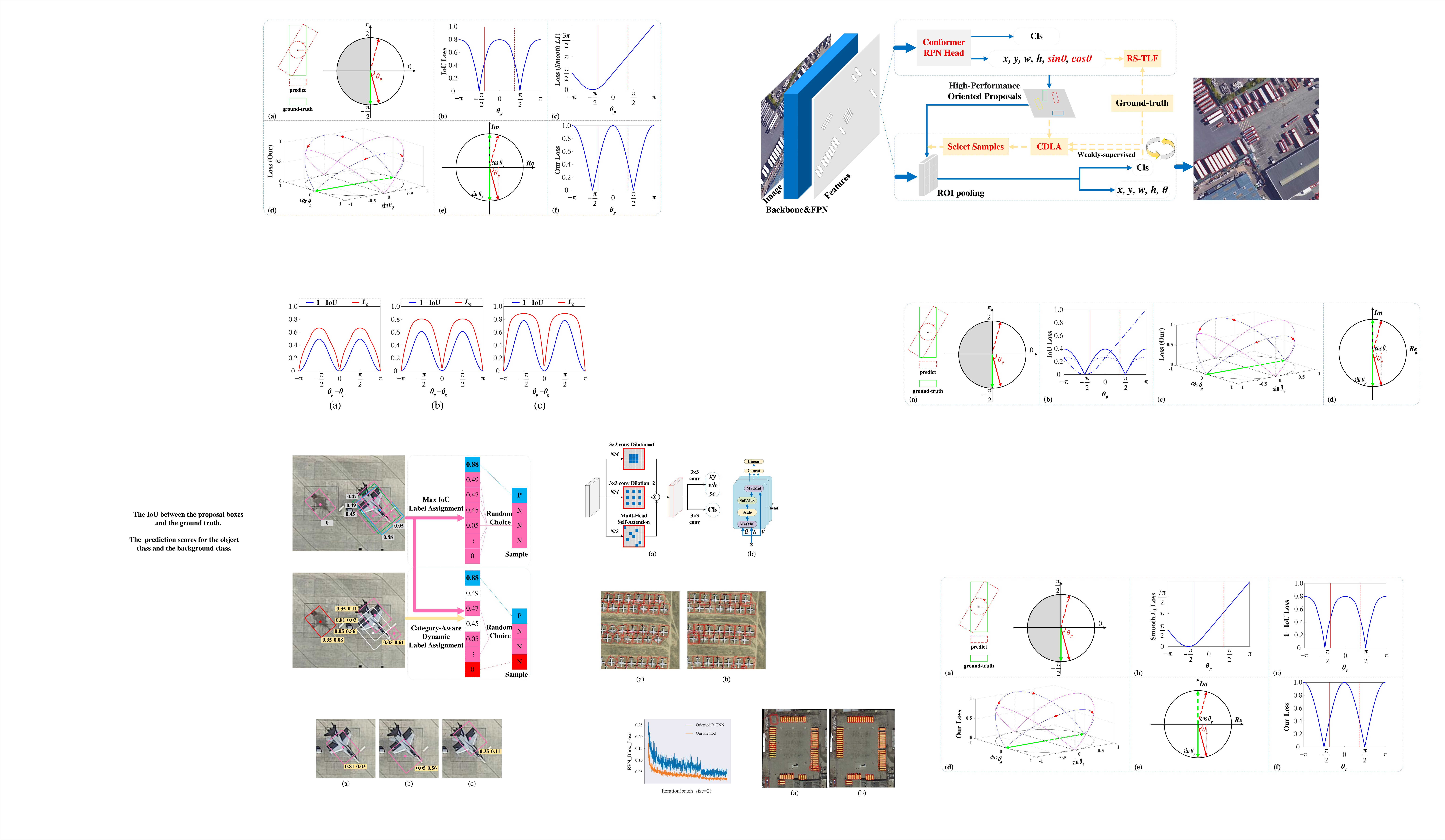}
	\caption{The RPN bounding box loss and iteration curves of the Oriented R-CNN and our.}
	\label{fig_9}
\end{figure}

\begin{figure}[!t]
	\centering
	\includegraphics[width=3.0in]{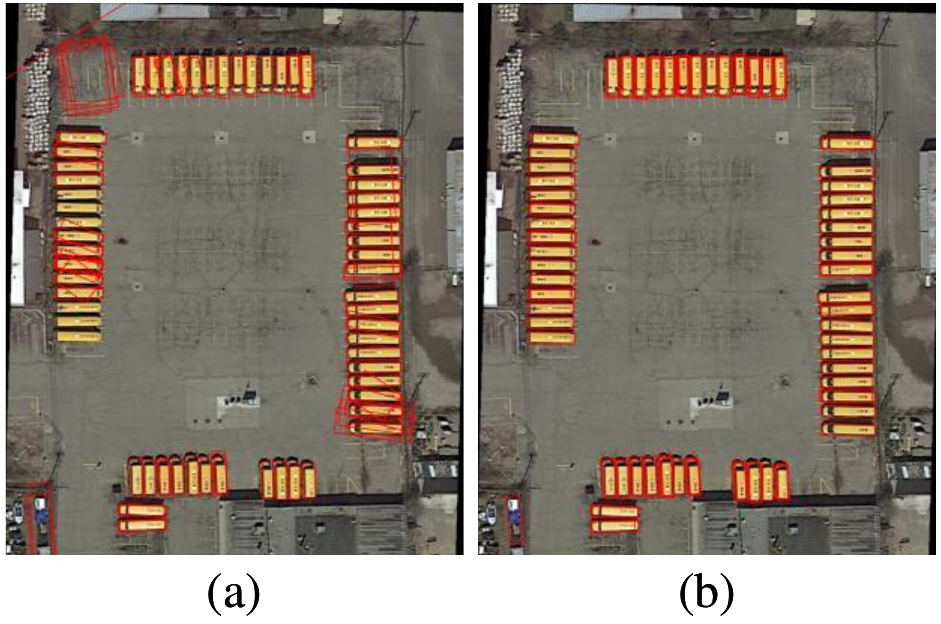}
	\caption{(a)The top 100 proposals generated by the RPN of Oriented R-CNN. (b)The top 100 proposals generated by our RPN. }
	\label{fig_10}
\end{figure}

The DOTA-v1.0 dataset was used for ablation studies. In these experiments, Faster R-CNN \cite{TPAMI2017_FasterR-CNNa} with ResNet50 \cite{CVPR2016_Resnet} as the backbone was employed as the baseline method. As shown in Table \ref{tab:table7}, the baseline method achieves an mAP of 73.40\%. By modifying the RPN with OBB, we established a new baseline, achieving an mAP of 74.35\%. We used a separate prior for each innovation point to ensure a fair comparison. The step-by-step improvement in mAP validated the effectiveness of each design. The three innovations proposed in this paper resulted in a total improvement of 2.84\% in mAP.

\subsubsection{Effectiveness of Loss}
As shown in Table \ref{tab:table7}, using the TLF module solely on the baseline improved the mAP value by 1.83\%, achieving an mAP of 76.18\%. We also demonstrate the bounding box loss and iteration of Oriented R-CNN and our method in Fig. \ref{fig_9}, showing that our loss function exhibits strong stability.

\subsubsection{Effectiveness of Conformer RPN Head}
The conformer RPN head module needs to be used based on the TLF. As shown in Table \ref{tab:table7}, the integration of the conformer RPN head module increases the mAP by 0.5\% in the third row compared to the first row, and by 0.56\% in the fifth row compared to the fourth row. The results indicate that the combination of convolution and multi-head self-attention helps in better capturing the correct classification information and sine-cosine components of the object angles. The conformer RPN head and TLF together contribute to generating high-quality proposals for ROI. Visualization comparison with Oriented R-CNN is shown in Fig. \ref{fig_10}.

\subsubsection{Effectiveness of CDLA}
As shown in Table \ref{tab:table7}, using only the CDLA module increased the mAP on the baseline by 1.99\%. Adding the CDLA module to the TLF resulted in a 0.45\% increase in mAP. Furthermore, incorporating the CDLA module on top of both the TLF and conformer RPN head increased the mAP by 0.51\%. The above results indicate that dynamically adjusting negative samples through category feedback facilitates flexible and reliable sample classification, which is more advantageous for network learning.

\section{CONCLUSION}
This study effectively addresses the issue of inconsistent parameter regression and boundary problems by designing TLF through angle regression analysis on the complex plane. This loss function offers sufficient flexibility to be integrated into any oriented detection framework. To better enable detectors to learn the complex plane coordinates of angles, a conformer RPN head is designed. Improvements in the loss function and consistent RPN header generate high-quality oriented proposals. To fully leverage high-quality proposals, a category-aware dynamic label assignment method based on predicted category feedback is proposed. Experimental results demonstrate that this work achieves highly competitive performance on four well-known remote sensing benchmark datasets.

\bibliographystyle{IEEEtran}
\bibliography{IEEEabrv,./TGRS.bib}

\end{document}